\def\year{2021}\relax
\documentclass[letterpaper]{article} % DO NOT CHANGE THIS
\usepackage{aaai21}  % DO NOT CHANGE THIS
\usepackage{times}  % DO NOT CHANGE THIS
\usepackage{helvet} % DO NOT CHANGE THIS
\usepackage{courier}  % DO NOT CHANGE THIS
\usepackage[hyphens]{url}  % DO NOT CHANGE THIS
\usepackage{graphicx} % DO NOT CHANGE THIS
\usepackage{natbib}  % DO NOT CHANGE THIS AND DO NOT ADD ANY OPTIONS TO IT
\usepackage{caption} % DO NOT CHANGE THIS AND DO NOT ADD ANY OPTIONS TO IT
\usepackage{microtype}

\usepackage{booktabs}
\usepackage{subfig}

\newcommand{\emoji}[1]{\includegraphics[height=1.40\fontcharht\font`\A]{emoji_images/#1.png}}

\title{
Identity Signals in Emoji do not Influence Perception of Factual Truth on Twitter}
\author {
    Alexander Robertson,
    Walid Magdy,
    Sharon Goldwater \\
}
\affiliations {
    School of Informatics, University of Edinburgh \\
    alexander.robertson@ed.ac.uk, wmagdy@inf.ed.ac.uk, sgwater@inf.ed.ac.uk
}

\begin{document}

\maketitle

\begin{abstract}

Prior work has shown that Twitter users use skin-toned emoji as an act of self-representation to express their racial/ethnic identity. We test whether this signal of identity can influence readers' perceptions about the content of a post containing that signal.
%the behaviour of those who see it, in terms of expressing bias for or against particular identities. 
In a large scale (n=944) pre-registered controlled experiment, we manipulate the presence of skin-toned emoji and profile photos in a task where readers rate obscure trivia facts (presented as tweets) as true or false. Using a Bayesian statistical analysis, we find that neither emoji nor profile photo has an effect on how readers rate these facts. This result will be of some comfort to anyone concerned about the manipulation of online users through the crafting of fake profiles.

\end{abstract}

\section{Introduction}

\begin{table*}[!ht]
\centering
\resizebox{\textwidth}{!}{%
\begin{tabular}{@{}llllcc@{}}
\toprule
Hyp. & Condition A                    & Condition B                   & Description                                              & BF (Black) & BF (White) \\ \midrule
H1a  & No photo, no emoji             & No photo, Black-aligned emoji & Sensitive to presence of emoji               & 0.109                          & 0.443                          \\
H1b  & No photo, no emoji             & Black photo, no emoji         & Sensitive to presence of photos              & 0.099                          & 0.087                          \\
H1c  & No photo, no emoji             & No photo, White-aligned emoji & Sensitive to presence of emoji               & 0.186                          & 0.073                          \\
H1d  & No photo, no emoji             & White photo, no emoji         & Sensitive to presence of photos              & 0.834                          & 0.142                          \\
H2a  & No photo, matching emoji       & No photo, non-matching emoji  & Preference for identity-matching emoji         & 0.098                          & 0.434                          \\
H2b  & Matching photo, no emoji       & Non-matching photo, no emoji  & Preference for identity-matching photos      & 0.284                          & 0.086                          \\
H3a  & Matching photo, matching emoji & No photo, matching emoji      & Emoji encode distinct identity signal & 0.085                          & 0.108                          \\
H3b  & Matching photo, matching emoji & Matching photo, no emoji      & Photos encode distinct identity signal & 0.084                          & 0.075                          \\ \bottomrule
\end{tabular}%
}
\caption{Overview of hypotheses tested, in terms of photo/emoji conditions. These were tested separately for Black and White readers: matching/non-matching means that the skin tone of the emoji was aligned with the self-reported ethnicity of the reader. Final columns show Bayes Factors for t-tests and are discussed in the analysis/discussion section.}
\label{tab:hypoth}
\end{table*}

As the influence of social media grows, it becomes ever more important to understand factors contributing to the spread of misinformation and ``fake news'', as well as how people's beliefs may be swayed online. In this paper, we investigate the role of author identity, as signalled through emoji. Can these identity signals influence the extent to which a reader accepts information as being true? The answer has important consequences not only for whether and how authors choose to signal their own identity, but also for studying the deliberate manipulation of identity signals in online content.

%We build on three distinct outcomes of prior work. First, people use emoji online that reflect their identity and infer author identity from the emoji they see. Second, linguistic signals can influence attitudes and perceptions of others. Third, knowledge of a person's identity can influence how we behave towards that person. Combining these findings, we examine the role of identity-related emoji in influencing behaviour. 

% Moved to Background
% Decades of social science research show that aspects of identity (i.e. socially meaningful categories such as gender, class, or ethnicity) can be signalled in many ways, and that these signals can influence our attitudes, perceptions, and behaviour towards a person \cite{campbell2010naturesociopercep}. Language is an important signal of identity and can affect one person's beliefs about another, such as whether they are trustworthy or intelligent \cite{lambert1960evaluational}. Identity signals can also affect behaviour: for example, using African American-sounding names on Uber resulted in increased wait times and more cancellations by drivers \cite{ge2016racial}. 
%Finally, there is some evidence that signals of a person's identity can influence beliefs about whether information they convey is true, \emph{even when they are only repeating information from others}. This last finding (described in more detail below) is particularly relevant to the spread of (mis)information on social media, and forms the basis for the work in this paper. We also build 

In this paper, we focus on the connection between identity signals and beliefs (rather than the behaviour that might result, such as retweeting). As the main identity signal of interest, we use skin-toned emoji, building on previous work showing that these are used as a form of self-representation (i.e., identity signalling)~\cite{robertson2018,robertson2020}, and that readers make use of these signals to infer aspects of the author's identity~\cite{robertson2021}. We hypothesise that identity signals in a tweet's emoji will affect the extent to which a reader believes the content of that tweet.

In a large scale (n=944) pre-registered controlled  experiment, we manipulate identity signals in tweets containing trivia facts and measure the extent to which readers perceive those facts to be true or false.
%
% Our experimental setup is based on the work of \citet{levari-credibility}, which looked at the impact of the difficulty of processing accents rather than identity, but we can easily adapt their approach to our needs. Instead of manipulating accent, we can manipulate signals of identity in profile photos and emoji. Through a large scale (n=944) carefully controlled pre-registered\footnote{Available at https://osf.io/a8r6q/} experiment, we measure the extent to which readers perceive social media content to be true or false. We manipulate expression of author ethnic identity in such content in two ways: the emoji present in a social media post and the user profile photo associated with a post. 
%
We find evidence against any effect on perceptions for either emoji or profile photo. Under a Bayesian statistical analysis, our experimental data supports the null hypothesis over the alternative in all of our analyses. This result is consistent across all experimental conditions, for two groups of participants based on self-reported ethnicity, and persists following adjustments to the experiment to address possible confounding factors in its design. Although these findings did not support our hypothesis, we argue that skin-toned emoji not affecting online behaviour in the way we expected is actually a positive result in the context of online manipulation, misinformation and fake news.

\section{Background}

Decades of sociology and linguistic research show that aspects of identity (i.e. socially meaningful categories such as gender, class, or ethnicity) can be signalled in many ways \cite{goffman1959presentation} and that these signals can influence our attitudes, perceptions, and behaviour towards a person \cite{campbell2010naturesociopercep,labov1972sociolinguistic}. Language is an important signal of identity and can affect one person's beliefs about another, such as whether they are trustworthy or intelligent \cite{lambert1960evaluational}. Identity signals can also affect behaviour: for example, using African American-sounding names on Uber resulted in increased wait times and more cancellations by drivers \cite{ge2016racial}.

% Emoji and identity
Emoji can also signal aspects of identity. For example, gender can be signalled consciously through choosing to use a male or female emoji (or neither) \cite{barbieri2018gender} or perhaps less consciously via the frequency of use of particular emoji such as \emoji{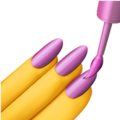} or \emoji{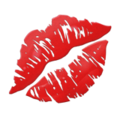} \cite{chen2018genderlens}. Here, we focus on ethnic identity, as signalled by skin-tone modifiers. Twitter users overwhelmingly use these to express their own identity \cite{robertson2018}, though referring to \emph{other} people in specific situations has been observed \cite{robertson2020}. Expression of identity by authors is only one side of the story --- readers of tweets containing light or dark toned emoji also associate authors with a congruent Black or White identity \cite{robertson2021}, and this association is separate from linguistic signals such as use of particular topics, phrases or syntax associated with Black or White speakers.

The connection between identity signals and behaviour has been studied in experimental economics, using the Trust Game. Pairs of anonymous players are designated sender and responder. The sender has a fixed amount of money and can keep the whole sum or send a portion to the responder. Any money sent this way will be tripled. The responder may keep everything they are sent, or return a portion back to the sender. Pairs play one round consisting of the sender making their decision, then the responder making their response. \citet{buchan2008trust} found an effect of gender on trust (as measured by the amounts sent/received), with women senders less likely overall to send money, but to send more to male responders than to female. \citet{stanley2011implicit} found a relation between the extent of a person's racial bias (as measured through the Implicit Attitudes Test (IAT) \cite{greenwald2002unified}) and the extent to which they trust players of a different racial background.

More recently, \citet{babinemojitrust2020} examined the effect of emoji when trust game players are allowed to communicate. Using emoji increased trust and reciprocation between players, but of particular relevance to our work here is the finding that use of skin-toned emoji may result in less trust when those skin tones are darker. \citeauthor{babinemojitrust2020} did not set out to evaluate the impact of skin tone and so this result must be regarded as preliminary. However, it motivates our work here as it is suggestive of skin-toned emoji being able to induce bias to the extent that behavioural outcomes are affected.

Our experimental setup is based on \citet{levari-credibility}, who looked at the effect of a speaker's accent on how we react to what they say. Participants listened to speakers with native, mild foreign or strong foreign accents reading trivia facts such as ``Ants never sleep''. On a scale from 0 to 100, participants rated statements from definitely false to definitely true. Accented statements were rated as less true even though participants were informed that the speakers were simply reciting statements written by the native-speaking experimenters. \citeauthor{levari-credibility} focused on processing difficulty as the explanatory factor for this outcome, rather than out-group or ethnic bias, but we would argue that the foreign accents used in the experiments are likely a highly salient indicator of race/ethnicity. To emulate the `reciting' aspect, we can use retweets.

\section{Hypotheses}

Before describing our experimental setup in detail, we will state the hypotheses it is intended to test and the rationale for these hypotheses. The experimental conditions arising from these hypotheses are shown in Table~\ref{tab:hypoth}.

First, we predict that the identity signal encoded in emoji \cite{robertson2021} will have an effect on how readers evaluate trivia statements in terms of being true or false, similar to the findings of \citet{levari-credibility}. We also assume that a profile photo also encodes an identity signal.

Second, we predict that readers will be sensitive to the kind of identity encoded in emoji and photos. Readers may prefer their own identity (similar to in-group bias in favour of own gender \cite{rudman2004gender}) or they may prefer another identity, for example if all users show a similar bias against a particular group
%similar to in-group bias against own race 
\cite{valla2018not}.

Third, we predict the presence of both a photo and an emoji signal will have a larger effect than either one alone, as per \citet{robertson2021} who found a ``boosting effect'' of an emoji and a language signal of identity when users were asked to guess the identity of the author.

We test our hypotheses separately with Black and White participants (self-identified; see Participants section). This is primarily to ensure we do not mistakenly extend any effects found in one group to all groups, as prior work has shown differences in how Black and White social media users use and perceive skin-toned emoji in relation to expressions of identity \cite{robertson2020,robertson2021}.

\section{Experimental setup}

Before proceeding, we acknowledge that some terms do not have universally agreed-upon definitions. Within the framework of Critical Race Theory \cite{crt-hci}, race and ethnicity are socially constructed. As such, the meaning of these and related terms depends on context, audience and the user. We use Black and White in our work to denote the self-identified ethnic identity of the participants. For emoji, we use Black-aligned and White-aligned to denote emoji which prior research \cite{robertson2018} shows to be preferred for self-representational usage by people of Black or White identities. Furthermore, we acknowledge that Black and White are not the only ethnicities which can be represented using emoji skin tones. However, we restrict our experimental setup to consider only these two options and make no claims as to how any findings may apply to other identities.

The experiment follows the setup of \citet{levari-credibility}, with the main difference that it was conducted online. Participants were shown 31 tweets containing trivia and informed that the trivia could be true or false. Their task was to mark on a slider from 0 to 100 whether they thought the trivia was \emph{definitely false} or \emph{definitely true}. Participants were explicitly asked not to look up any trivia and informed that the correct answers would be shown upon completion of the study.

\paragraph{Stimuli}

\begin{figure}
\centering
\includegraphics[width=\columnwidth]{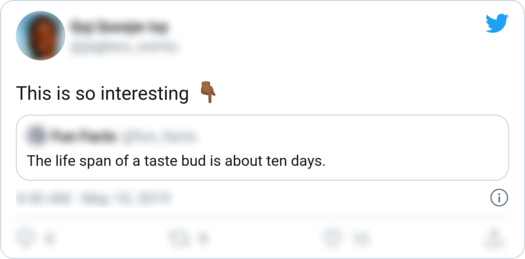}
\caption{Example of a critical trial from one block, showing manipulation of the identity encoded in profile photos and emoji.}\label{figure:stimuli}
\end{figure}

Critical trials manipulated the identity signals present in emoji and photos, as shown in Figure~\ref{figure:stimuli}. Emoji were either dark-toned (aligned with Black identities), light-toned (aligned with White identities) or absent. Only the downwards pointing finger \emoji{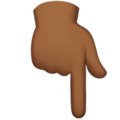} \emoji{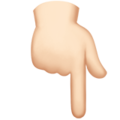} was used. Photos signalled a probable Black identity, White identity or no obvious identity (e.g. a nature photo). There were 7 critical combinations of emoji/photo. Photos were female faces generated by StyleGAN\footnote{https://github.com/NVlabs/stylegan}, to avoid using photos of real people, and blurred to obscure all detail except skin tone. In a pre-norming study, 25 Black and 25 White participants guessed the ethnicity of the blurred faces. We retained only those which the majority ($\geq$95\%) of participants considered to show a Black or a White person.

All trivia in the critical trials were true, but were likely to be neither well-known to be true nor generally thought (mistakenly) to be false. This was to prevent any ceiling effect in the ratings of the stimuli and was determined through a pre-norming study of trivia facts taken from Snapple soft drink lids.\footnote{See https://www.snapple.com/real-facts for a full list} 20 Black and 20 White participants rated 100 random facts as either true or false. We randomly sampled 7 from those facts where participants were at chance in deciding between true and false. Mean ratings for these facts were 47.5 (stdev 0.0932) for Black participants, 44.6 (stdev 0.0935) for White participants. 

To control for any possible impact of individual facts on participant judgements, each of these facts appeared in each critical combination of emoji/photo. This was done in blocks, such that participants only saw each fact once in a single emoji/photo combination, but across the entire experiment all critical facts were judged in all combinations.

Filler trials (to obscure the intent of the experiment) were balanced across two types of fact. Twelve were facts that pre-norming participants generally considered to be true/false (6 of each), and twelve were facts we created which were very obviously true or very obviously false (again, 6 of each). In addition, profile photos could be colourful and non-human (e.g. scenery) and blurred verified account ticks could be present in user names. Also, a wider range of emoji, including skin-toned ones, could appear in the main retweeter's text.

All stimuli were presented as a main user retweeting another account, with some comment expressing surprise or incredulity. Authentic-looking tweets were automatically generated from a copy of the Twitter HTML/CSS and saved as PNG files, to discourage participants looking for answers online by copying/pasting the text. The surprise/incredulity comments were selected based on a pre-norming study. 50 Black and 50 White participants provided three examples of how they would comment if sharing such a tweet. The most common responses shared between Black and White participants were chosen and sampled from at random.

Display names, user names, dates, number of likes/retweets/comments were all randomised before being blurred. In critical trials, the account being retweeted was always unverified and had a default grey display photo. In filler trials, these properties were randomly assigned. Finally, trials were presented in randomised order per participant. All stimuli can be found in our Open Science Framework pre-registration.\footnote{See https://osf.io/a8r6q/}

\paragraph{Participants}

Participants were recruited through Prolific.co on the basis that they are native speakers of English, US citizens resident in the US and self-identified their ethnicity as either Black or White. We did not request or record any non-pertinent personal information (e.g. age, gender), in accordance with the University's data handling protocols. In all pre-norming and experimental tasks, participants were paid at a rate equivalent to the National Living Wage of £9.50 per hour. Approval for the study was obtained in advance from the University's ethics board. In total, 944 participants (472 Black, 472 White) were recruited for the main study, to ensure sufficient power to determine with a good degree of confidence whether our null or alternative hypothesis is true (using a Bayesian statistical analysis, described below). Power was determined using the SSDbain package in R \cite{fu2020sample}.

\section{Analysis plan}

The experiment generates a sequence of true/false ratings (from 0 to 100) per critical combination of emoji/photo. We analyse the means of these ratings for differences using a Bayesian two-tailed independent samples t-test \cite{van2020jasp}, as implemented by the JASP\footnote{https://jasp-stats.org/} software package. The Bayesian t-test is parameterised by a prior distribution over our assumptions of the effect size. We follow standard practice of using a Cauchy prior with mean 0 and scale 0.707 \cite{quintana2018bayesian}.

We chose a Bayesian version of the t-test for comparing sample means as it allows us to report Bayes Factors rather than p-values, which means we can quantify the amount of support for or against our hypotheses, rather than simply testing whether the null hypothesis can be rejected~\cite{rouder2009bayesian}. The Bayes Factor (BF) is the ratio of the marginal likelihoods of two competing hypotheses (H0 and H1), given some observed data. Multiplying the prior odds of H0 and H1 (i.e. the ratio of P(H0) to P(H1)) by the BF gives the posterior odds: the ratio of P(H0$\vert$data) to P(H1$\vert$data). The BF quantifies the extent to which the observed data supports the hypotheses under consideration. For our purposes, the two competing hypotheses are no difference in means (H0) and a significant difference in means (H1). 
%An additional benefit of this approach is that it allows us to speak about support for either hypothesis, rather than only in terms of rejecting the null hypothesis \cite{rouder2009bayesian}.
A BF value of 1 means the observed data supports each hypothesis equally well (i.e. it is uninformative), and normally indicates additional data is needed or hypotheses need to be reformulated. A BF value greater than 1 means there is more support for H1, while a value less than 1 means more support for H0.

\section{Results}

\begin{figure}
\centering
\includegraphics[width=\columnwidth]{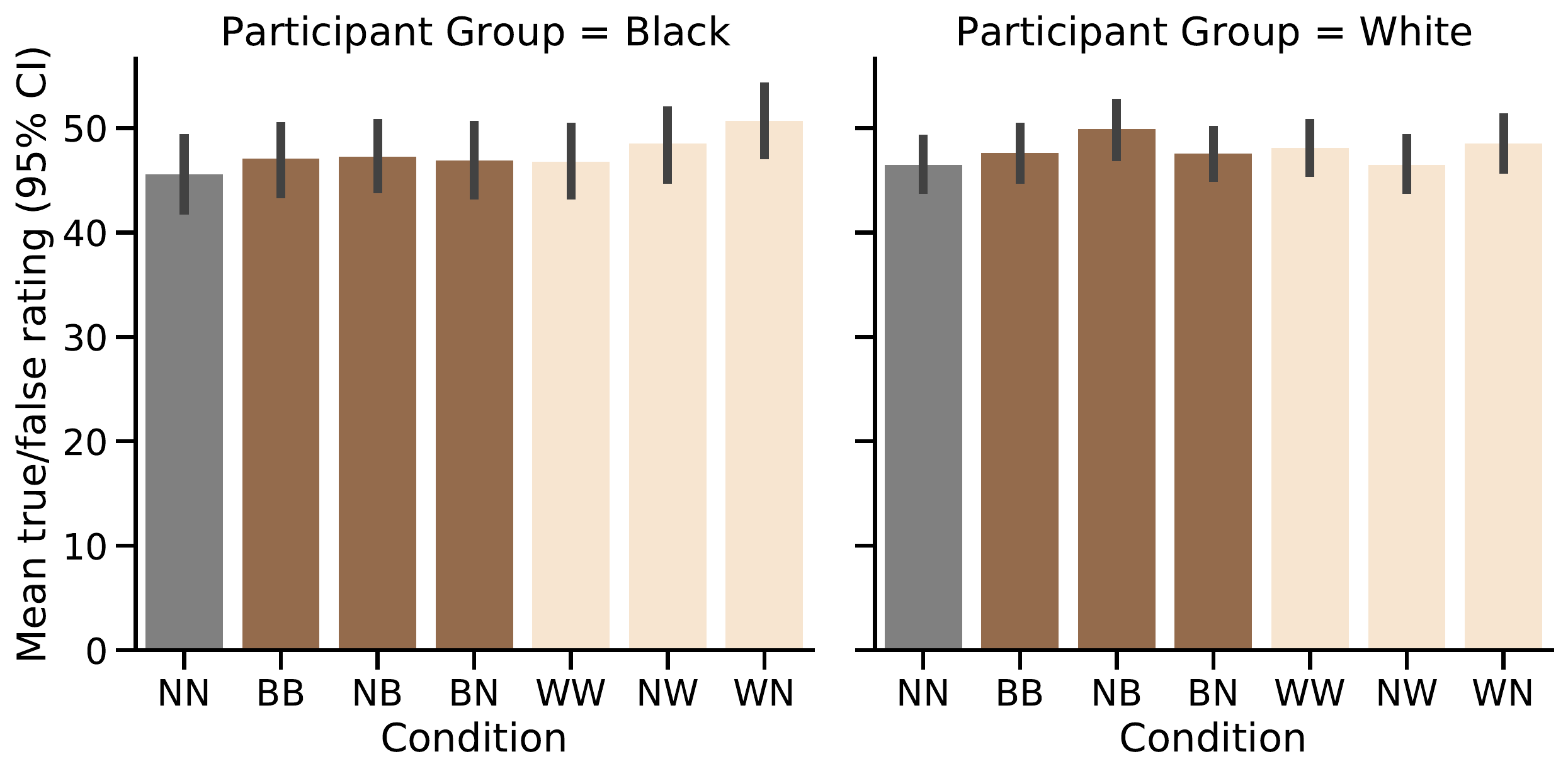} 
\caption{Mean true/false ratings (with 95\% confidence intervals) per participant group. Each bar represents an experimental condition, with labels XY representing Emoji status (Black, White, No emoji) and Photo status (Black, White, No photo). Bars are coloured based on their prevalent identity signal (grey = no signal, brown = Black signal, beige = White signal)}\label{figure:results}
\end{figure}

Mean true/false ratings per experimental condition for each participant group are shown in Figure~\ref{figure:results}. The results of all analyses for both participant groups were in favour of H0 --- there was no effect on the mean true/false ratings in any subsets of ratings selected for comparison. Table~\ref{tab:hypoth} reports the results of all t-tests in terms of Bayes Factor, from which it can be seen that experimental evidence offers between 1.2 and 13.4 times more support (i.e. 1/BF) for H0 in all cases than for the alternative hypothesis of a difference in mean true/false ratings.
From this, we conclude that neither emoji nor profile photo, either alone or together, influence a reader's behaviour in the true/false trivia task.

\section{Further Analysis and Discussion}

Although our experiments were motivated by a range of prior work, used carefully designed and normed stimuli, and involved almost 1000 participants, the evidence we found against any effect of identity signals prompted us to reflect on the experimental design and explore possible issues.

% REMOVE THIS if need to make space.
\begin{figure}
\centering
\includegraphics[width=\columnwidth]{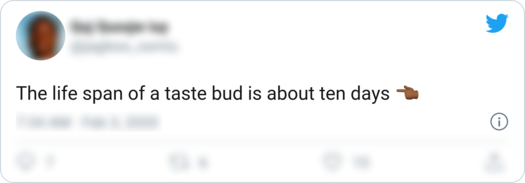} 
\caption{Example of a critical trial from one block, with the user directly tweeting the trivia, rather than retweeting another account.}\label{fig:noquote}
\end{figure}

Did the presentation of the fact as a retweet of an account with no identity signals cause readers to focus only on the retweeted account? We ran a small scale version (n=80, White participants only) of the experiment without the retweet (Figure \ref{fig:noquote}) and again obtained statistical support only for the null hypothesis (BF range of 0.181-0.783, 1.2 to 5.5 times more support for H0).

Was participants' familiarity with Twitter a factor? The Prolific.co platform provided data on participants' self-reported Twitter use and we separated groups into Twitter users and non-users. Again, we obtained qualitatively the same results, regardless of whether stimuli were tweets or retweets (BF range of 0.144-0.385, 2.6 to 6.9 times more support for H0).

It may be that profile photos had no effect on behaviour because they were blurred. We had normed these blurred photos to ensure the skin tone was determinable, but this alone may not be sufficient to cause any bias or preferential behaviour in this task. As emoji were not blurred but also had no effect, we suspect it is unlikely that unblurring profile photos would have any effect.

Perhaps we should have included a measure of bias by administering an IAT, as per \citet{stanley2011implicit}, and balanced participant numbers not only in terms of self-reported racial identity but also in terms of pre-existing bias. This we leave for future work and might give an interesting perspective on whether emoji can trigger biases or preferences which are already known to exist, but see prior work on issues with failure to IAT results and poor correlation between IAT scores and behavioural outcomes \cite{blanton2009strong}.

Finally, the true/false trivia task may itself not be best suited to detecting the effect we set out to find. %Follow-up work could use a form of the Trust Game from economics specifically adapted to include signals of identity such as skin-toned emoji and profile photos.
These facts are by definition trivial, and whether they are true or false is likely unimportant to the reader. By contrast, facts with real-world consequences or emotional valence might be more subject to unconscious biases surrounding the identity of the source. We did not explore that type of fact here, since the experiment would be far more challenging to design, raising both ethical questions (we want to avoid accidentally spreading misinformation) and design issues (it is difficult to come up with plausible stimuli). However, such an experiment could provide important further evidence regarding situations in which identity does or does not affect perceptions of truth.

%\section{Discussion}

How any findings from such work would translate to the case of Twitter is not clear. On the positive side, we found
%One positive outcome of this study is that there is 
strong evidence that the participants in our study were not swayed by identity signals to rate trivia as more or less true/false. This has consequences for work in disinformation. In an analysis of tweets from a Russian government-funded troll farm known as the Internet Research Agency, \citet{freelon2020black} found that fake accounts included aspects of racial presentation, using Black profile photos and names associated with Black Americans. The practice of Black impersonation in this case appeared to result in increased engagement with the content from real Twitter users, in terms of likes and retweets. However, as \citeauthor{freelon2020black} note, there is no data on the identities of the users engaging with those tweets. One way of reconciling the findings of \citeauthor{freelon2020black} and ours is to suppose that the increased engagement came only from additional fake bot accounts (though not necessarily from the same troll farm, as the authors found little evidence of self-amplification) rather than real users. The real users in our study seem indifferent to such identity signals in the context of behavioural influence. To better understand the interaction of users, identity, behaviour and disinformation, future work in this area should combine the experimental approach of our study and the content perspective of \citet{freelon2020black}.

\section{Conclusion}

In a controlled experiment, we found that identity signals in emoji or photos do not influence the extent to which readers perceive information as true or false. This is potentially good news in light of attempts to manipulate social media users, but future work should examine whether this result obtains for different, more realistic, types of information. Our methodology is well-documented due to the pre-registration process, making such work easier for interested researchers.
%so such work can easily be done by any interested party.

\bibliography{bibfile}

\end{document}